\documentclass[conference]{IEEEtran}
\IEEEoverridecommandlockouts
\usepackage{amsmath,amssymb,amsfonts}
\usepackage{algorithm}
\usepackage{algorithmic}
\usepackage{graphicx}
\usepackage{textcomp}
\usepackage{xcolor}
\usepackage[T1]{fontenc}
\usepackage{cite}
\usepackage{float}
\usepackage{placeins}
\usepackage{float}
\usepackage{placeins}
\usepackage{multicol}
\usepackage{cuted}
\usepackage{caption}
\usepackage{xcolor}
\usepackage{soul}
\usepackage{fancyhdr}
\usepackage[colorlinks=true, linkcolor=blue, citecolor=blue, urlcolor=blue]{hyperref}

\def\BibTeX{{\rm B\kern-.05em{\sc i\kern-.025em b}\kern-.08em
    T\kern-.1667em\lower.7ex\hbox{E}\kern-.125emX}}
\begin{document}

\title{Fine-Grained Cat Breed Recognition with Global Context Vision Transformer}

\author{
\IEEEauthorblockN{
Mowmita Parvin Hera$^{1}$, Md. Shahriar Mahmud Kallol$^{1}$, Shohanur Rahman Nirob$^{1}$,\\
Md. Badsha Bulbul$^{1}$, Jubayer Ahmed$^{1}$, M. Zohurul Islam$^{1}$,\\
Hazrat Ali$^{2}$ (Senior Member, IEEE), Mohammad Farhad Bulbul$^{1,*}$
}
\IEEEauthorblockA{$^{1}$Jashore University of Science and Technology, Bangladesh}
\IEEEauthorblockA{$^{2}$University of Stirling, Stirling, United Kingdom}
\IEEEauthorblockA{*Corresponding author: farhad@just.edu.bd}
}
\maketitle

\maketitle

\begin{strip}
\centering
\includegraphics[width=1\linewidth]{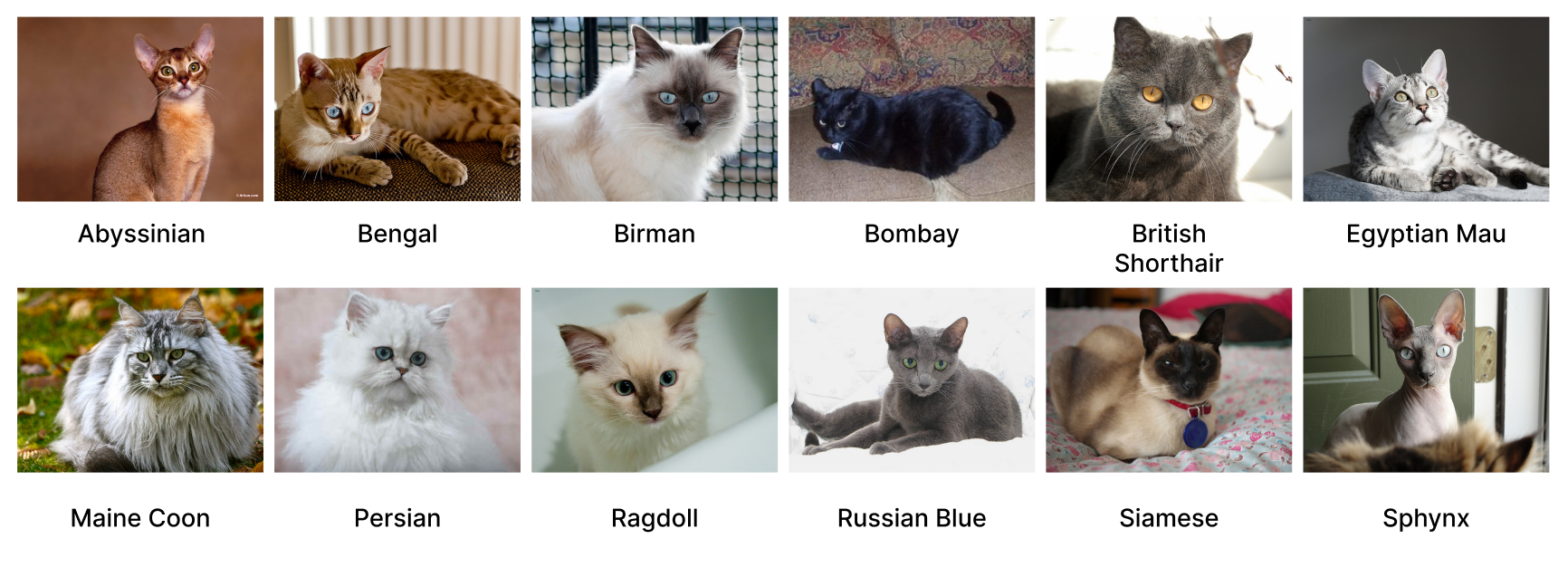}
\captionof{figure}{Sample images of cat breeds}
\end{strip}

\begin{abstract}
Accurate identification of cat breeds from images is a challenging task due to subtle differences in fur patterns, facial structure, and color. In this paper, we present a deep learning-based approach for classifying cat breeds using a subset of the Oxford-IIIT Pet Dataset, which contains high-resolution images of various domestic breeds. We employed the Global Context Vision Transformer (GCViT) architecture-tiny for cat breed recognition. To improve model generalization, we used extensive data augmentation, including rotation, horizontal flipping, and brightness adjustment. Experimental results show that the GCViT-Tiny model achieved a test accuracy of 92.00\% and validation accuracy of 94.54\%. These findings highlight the effectiveness of transformer-based architectures for fine-grained image classification tasks. Potential applications include veterinary diagnostics, animal shelter management, and mobile-based breed recognition systems. We also provide a hugging face demo at https://huggingface.co/spaces/bfarhad/cat-breed-classifier.
\end{abstract}

\begin{IEEEkeywords}
Cat breed classification, Deep learning, Oxford-IIIT Pet Dataset, Vision transformer, Fine-grained image classification.
\end{IEEEkeywords}

\section{Introduction}
\label{sec:introduction}

Cat breed classification is a challenging fine-grained visual recognition task with important applications in veterinary diagnostics, pet adoption systems, and animal welfare management. As a specialized case of fine-grained visual classification (FGVC), it requires distinguishing between classes with subtle inter-class differences while handling significant intra-class variation~\cite{b4}. Existing computational methods often struggle with high inter-class similarity (e.g., Ragdoll vs. Birman cats~\cite{b4}) and intra-class variability due to pose, lighting, and occlusions~\cite{b4}.

Previous studies have approached cat breed classification using techniques ranging from handcrafted feature extraction to advanced transformer-based models. Early FGVC methods relied on descriptors such as Scale-Invariant Feature Transform (SIFT), Local Binary Patterns (LBP), and Histogram of Oriented Gradients (HOG), combined with traditional classifiers like Support Vector Machines (SVMs) or k-Nearest Neighbors (k-NN)~\cite{b2}. For example, Parkhi et al.~\cite{b2} introduced a hybrid model for fine-grained cat breed classification that integrated shape and appearance information. Shape was captured through face detection using a deformable part model with HOG filters, while appearance was represented by a bag-of-words model based on dense SIFT features and pooling in spatial layout. The model also included an automatic segmentation step utilizing color and texture cues to separate pets from the background. Classification was performed using a multi-kernel support vector machine that integrated the scores for shape and appearance. The method achieved a cat breed classification accuracy of 66.07\% on the newly proposed Oxford-IIIT pet dataset, setting an important benchmark for research before the deep learning era. While effective at capturing local texture and shape information, these approaches required extensive manual feature engineering and showed limited robustness to variations in pose, lighting, and background~\cite{b4}.

Deep learning-based approaches using convolutional neural networks (CNNs) have since become the dominant approach for this task. Thus, several studies have investigated deep learning models for cat breed identification. Fawwaz et al.~\cite{fawwaz2021klasifikasi} compared architectures including VGG16, InceptionV3, ResNet50, and Xception, with Xception demonstrating the best performance on the Oxford-IIIT Pet Dataset.
Cahyo et al.~\cite{cahyo2023transfer} used the Xception model and 2400 images from 12 breeds in the Oxford-IIIT pet dataset to study the effects of transfer learning and fine-tuning on CNNs for cat breed classification. Their results showed that applying fine-tuning on top of transfer learning significantly improved the model's classification performance compared to using transfer learning alone. Ang et al.~\cite {ang2024petvision} proposed PetVision, a lightweight hybrid model built on MobileViTv3. This model integrated a coordinate attention mechanism with an improved inverse residual module, achieving a  75\% validation set accuracy. This research also indicated that image classification tasks are shifting from traditional, computationally intensive architectures to more practically applicable and efficient hybrid vision transformer architectures.

Vision Transformers (ViTs) offer a promising alternative by employing self-attention to capture long-range dependencies across the entire image. Dosovitskiy et al.~\cite{b10} first demonstrated the capability of ViTs to match or surpass CNN performance on large-scale classification benchmarks. Liu et al.~\cite{b9} extended this concept with the Swin Transformer, introducing hierarchical attention to improve scalability and efficiency. More recently, Hatamizadeh et al.~\cite{hatamizadeh2023global} developed the GCViT, which enhances global feature modeling, making it particularly suitable for fine-grained classification tasks.

Despite these advances, limited research has explored transformer-based architectures for cat breed classification. Existing studies either focus on CNNs~\cite{fawwaz2021klasifikasi} or hybrid architectures~\cite {ang2024petvision}, leaving the potential of global-context modeling underutilized. In this work, we address this gap by fine-tuning GCViT-Tiny~\cite{hatamizadeh2023global} on the Oxford-IIIT Pet Dataset~\cite{b2}. 

The remainder of this paper is organized as follows: %Section~\ref{sec:related} reviews related work in detail,
Section~\ref{sec:methodology} describes our proposed methodology, Section~\ref{sec:results} presents experimental results and comparisons, and Section~\ref{sec:conclusion} concludes with future research directions.

\section{Proposed Approach}
\label{sec:methodology}
This section outlines the proposed methodology, detailing the data preparation procedures and the architecture of the classification model employed.
\subsection{Pipeline Overview}
Our cat breed classification pipeline using GCViT ~\cite{hatamizadeh2023global} consists of sequential stages, starting with data collection and preprocessing of the Oxford-IIIT Pet dataset ~\cite{b2}. The model is then initialized with pre-trained GCViT weights, and images are prepared as patch embeddings with convolutional features. These embeddings pass through a transformer encoder with global context integration, and the refined \texttt{[CLS]} token is classified into breed categories. Note that the CLS token is a special learnable classification token that accumulates global image information by attention mechanisms. Its ultimate embedding is used as the global image representation for final classification. However, training optimizes model parameters, and final performance is evaluated using standard classification metrics.

\begin{figure}[!ht]
    \centering
    \includegraphics[width=1\linewidth]{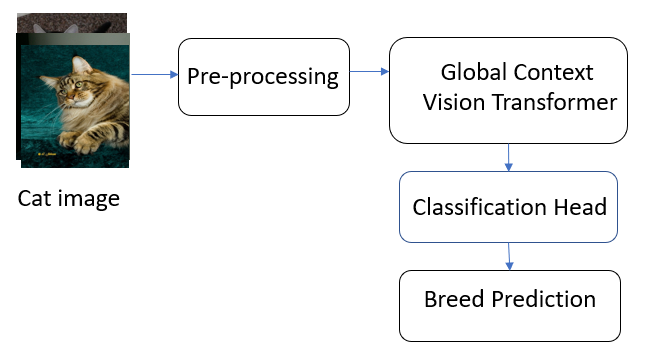}
    \caption{Cat breed classification pipeline}
    \label{fig:pipelineoverview}
\end{figure}

Fig. \ref{fig:pipelineoverview} presents the overall pipeline of the GCViT-based cat breed classification. In this section, we present a detailed description of these stages, organized into three main components: data preparation, the GCViT architecture, and training with breed prediction.

\subsection{Data Preparation}
For the training data, we apply a series of augmentation and preprocessing steps to improve the generalization ability of the model. Each image is randomly cropped and resized to $224 \times 224$ pixels within a scale range of $[0.8, 1.0]$, ensuring that variations in object scale and position are preserved. To account for natural pose differences, we apply random horizontal flipping and random rotations of up to $20^\circ$. In addition, color jittering is introduced by randomly varying the brightness, contrast, and saturation by up to $20\%$, which helps the model remain robust against illumination and color variations. After these augmentations, the images are converted into tensors and normalized using the mean and standard deviation values of the ImageNet dataset $([0.485, 0.456, 0.406], [0.229, 0.224, 0.225])$. This standardization ensures that the inputs have zero mean and unit variance across channels, facilitating stable and efficient training.  

The Oxford-IIIT Pet dataset ~\cite{b2} dataset comprises 12 classes with a total of 2,371 images. The training data is further split into training and validation subsets using an 80:20 ratio, resulting in 950 training samples and 238 validation samples. The test set contains 1,183 samples. To ensure balanced class representation across all subsets and provide a fair evaluation of the model’s performance, stratified sampling is employed.

\subsection{Classification Methodology}
\textbf{Patch and Convolutional Embedding:}  
The input image \( \mathbf{I} \) first passes through convolutional stem layers to extract local spatial features before patchification. The resulting feature map is divided into \( N \) non-overlapping patches:
\[
N = \frac{H}{P} \times \frac{W}{P}
\]
where \( P \) is the patch size (e.g., 16). Each patch \( \mathbf{x}_i \) is flattened and projected to a \(D\)-dimensional embedding:
\[
\mathbf{z}_i^0 = \mathbf{E} \mathbf{x}_i + \mathbf{p}_i
\]
where \( \mathbf{E} \) is the learnable projection matrix and \( \mathbf{p}_i \) is the positional encoding.

\textbf{Global Context Attention Block:}  
In contrast to standard self-attention, GCViT employs a Global Context Module (GCM) that incorporates global feature statistics into the attention mechanism. Given \( \mathbf{X} \in \mathbb{R}^{(N+1) \times D} \), queries, keys, and values are computed as:
\[
\mathbf{Q} = \mathbf{X} \mathbf{W}^Q, \quad
\mathbf{K} = \mathbf{X} \mathbf{W}^K, \quad
\mathbf{V} = \mathbf{X} \mathbf{W}^V
\]
The global context vector \( \mathbf{g} \in \mathbb{R}^D \) is obtained via a learnable aggregation function:
\[
\mathbf{g} = \mathcal{G}(\mathbf{X}) = \sum_{i=1}^N \alpha_i \mathbf{x}_i, \quad \alpha_i = \frac{\exp(\mathbf{w}_g^\top \mathbf{x}_i)}{\sum_{j=1}^N \exp(\mathbf{w}_g^\top \mathbf{x}_j)}
\]
This context is then fused into keys and values:
\[
\tilde{\mathbf{K}} = \mathbf{K} + \mathbf{g} \mathbf{W}^G_K, \quad
\tilde{\mathbf{V}} = \mathbf{V} + \mathbf{g} \mathbf{W}^G_V
\]
Scaled dot-product attention is computed as:
\[
\mathrm{Attention}(\mathbf{Q}, \tilde{\mathbf{K}}, \tilde{\mathbf{V}}) = \mathrm{softmax} \left( \frac{\mathbf{Q} \tilde{\mathbf{K}}^\top}{\sqrt{d_k}} \right) \tilde{\mathbf{V}}
\]

\textbf{Hierarchical Stages:}  
GCViT processes features through multiple stages with decreasing spatial resolution and increasing embedding dimension, combining convolutional downsampling with global context attention.

\subsubsection*{Training and Breed Prediction}
The final \texttt{[CLS]} token embedding is fed into a fully connected softmax layer:
\[
\hat{\mathbf{y}} = \mathrm{softmax}(\mathbf{W}_c \mathbf{z}_{\text{cls}}^L + \mathbf{b}_c)
\]
where \( C \) is the number of cat breeds. Training is performed using categorical cross-entropy loss with label smoothing (\( \epsilon = 0.1 \)) and the \textit{AdamW} optimizer. A cosine annealing learning rate schedule is used:
\[
\eta(t) = \eta_{\min} + \frac{1}{2} (\eta_{\max} - \eta_{\min}) \left( 1 + \cos\left( \frac{t\pi}{T} \right) \right)
\]

\section{Results and Discussion}
\label{sec:results}
In this study, experiments were conducted on a system equipped with an Intel Core i5-7500 CPU (3.40 GHz), 16 GB of RAM, and an NVIDIA GeForce GTX 1060 GPU with 6 GB of VRAM, running Windows 10. All the implementations were done using the PyTorch deep learning framework.

The pretrained GCViT tiny variant ~\cite{hatamizadeh2023global} was fine-tuned in a filtered cat subset of the Oxford-IIIT Pet Dataset~\cite{b2} to specialize the model for cat breed classification. We assessed the performance of the model on clean (unaltered) images to establish a baseline and demonstrate its capability to use global contextual cues for fine-grained classification.

 The pretrained GCViT model was fine-tuned on the training set and evaluated on the validation set, with the best checkpoint selected based on the highest validation accuracy. This optimal model was subsequently applied to the test set to assess its performance. Standard classification metrics i.e., accuracy, precision, recall, and F1-score were employed to evaluate the model. Although the maximum number of training epochs was set to 100, early stopping terminated training after 10 epochs to prevent overfitting. Since the test set comprised entirely unseen samples, the consistently high accuracy observed indicates strong generalization performance over just 10 epochs of training.

\begin{figure}[!ht]
\centering
\includegraphics[width=1\linewidth]{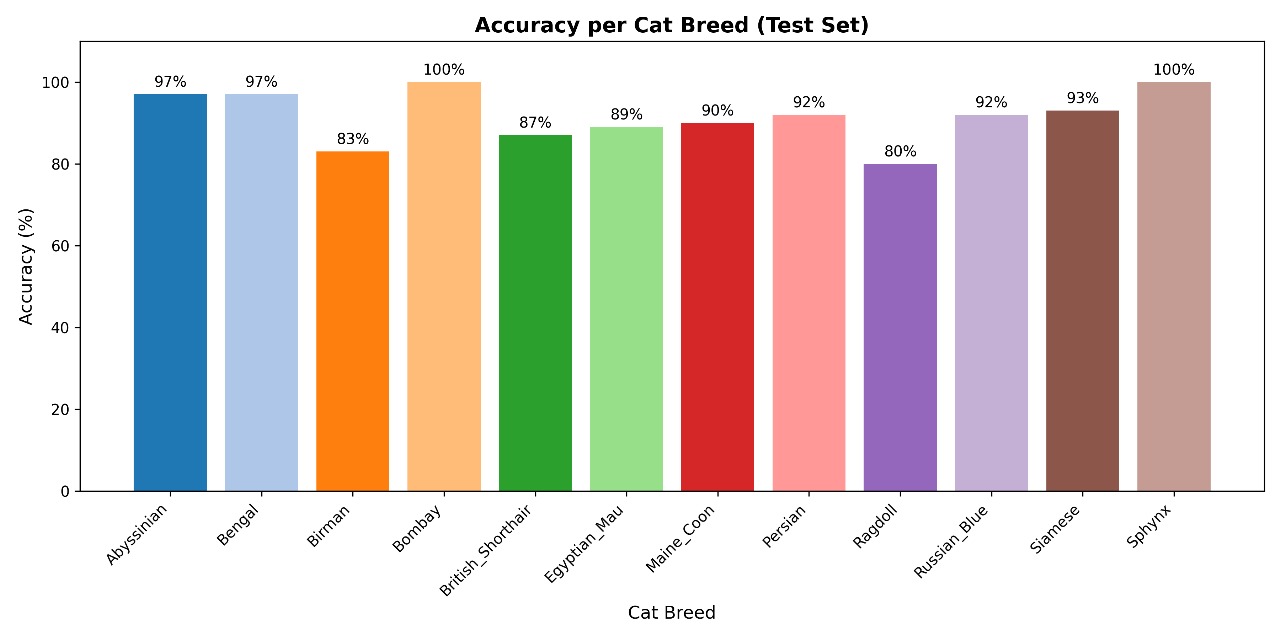}
\caption{Accuracy per cat breed on the test set}
\label{fig:accuracy_per_breed}
\end{figure}

\begin{figure}[H]
\centering
 \includegraphics[width=1\linewidth]{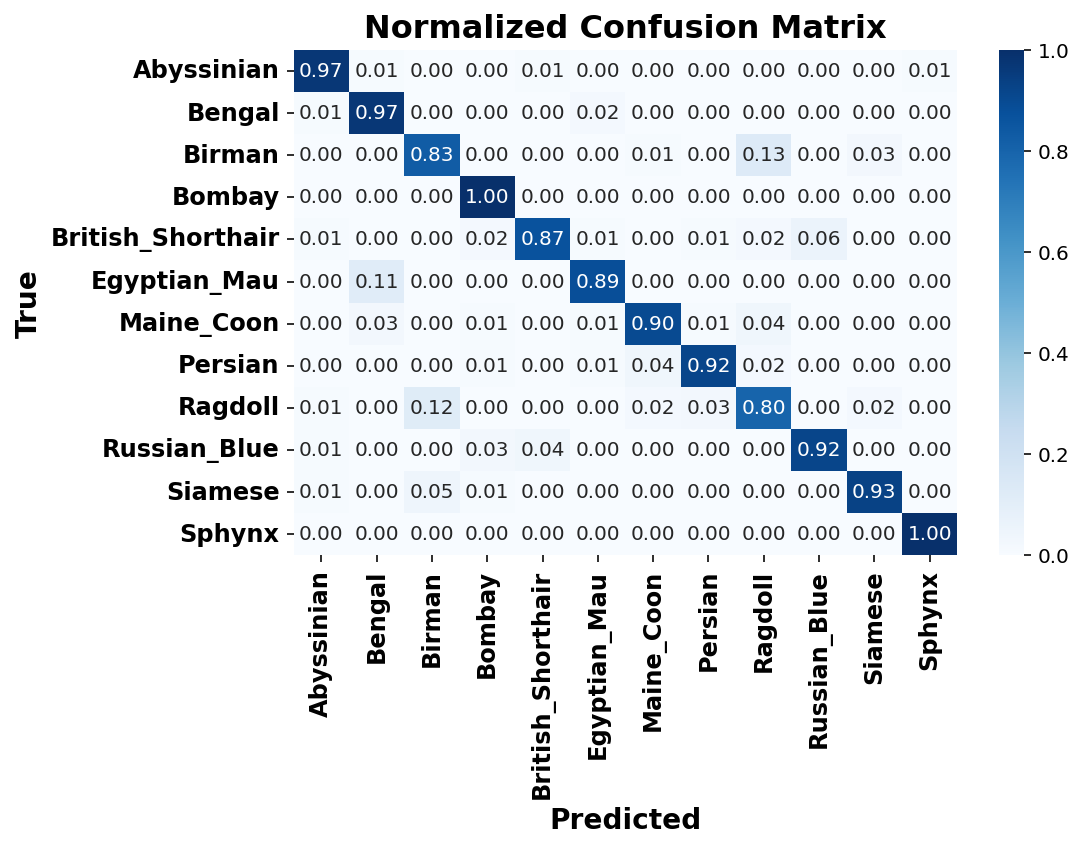}
\caption{Confusion matrix on the test set}
\label{fig:conf_matrix}
\end{figure}

Table~\ref{tab:classification_report_val} presents the detailed classification report, showing high performance for most breeds out of 12 breeds. The confusion matrix in Fig.~\ref{fig:conf_matrix} highlights minimal confusion between visually similar breeds.

\begin{table}[ht]
\centering
\caption{Classification report on the test set}
\label{tab:classification_report_val}
\renewcommand{\arraystretch}{1.2}
\begin{tabular}{|l|c|c|c|c|}
\hline
\textbf{Breed} & \textbf{Precision} & \textbf{Recall} & \textbf{F1-score} & \textbf{Support} \\
\hline
Abyssinian        & 0.95 & 0.97 & 0.96 & 98 \\
\hline
Bengal            & 0.87 & 0.97 & 0.92 & 100 \\
\hline
Birman            & 0.83 & 0.83 & 0.83 & 100 \\
\hline
Bombay            & 0.92 & 1.00 & 0.96 & 88 \\
\hline
British Shorthair & 0.95 & 0.87 & 0.91 & 100 \\
\hline
Egyptian Mau      & 0.95 & 0.89 & 0.91 & 97 \\
\hline
Maine Coon        & 0.93 & 0.90 & 0.91 & 100 \\
\hline
Persian           & 0.95 & 0.92 & 0.93 & 100 \\
\hline
Ragdoll           & 0.79 & 0.80 & 0.80 & 100 \\
\hline
Russian Blue      & 0.94 & 0.92 & 0.93 & 100 \\
\hline-
Siamese           & 0.95 & 0.93 & 0.94 & 100 \\
\hline
Sphynx            & 0.99 & 1.00 & 1.00 & 100 \\
\hline
\textbf{Accuracy}      &       &       & \textbf{0.92} & \textbf{1183} \\
\hline
\textbf{Macro Avg}     & \textbf{0.92} & \textbf{0.92} & \textbf{0.92} & \textbf{1183} \\
\hline
\textbf{Weighted Avg}  & \textbf{0.92} & \textbf{0.92} & \textbf{0.92} & \textbf{1183} \\
\hline
\end{tabular}
\end{table}

Fig~\ref{fig:accuracy_per_breed} illustrates the classification accuracy for each cat breed on the test set. The model demonstrates high and consistent accuracy across most categories, highlighting its strong generalization capability. Notably, the \textit{Sphynx} breed achieves the highest accuracy at \textbf{99\%}, indicating perfect classification performance for this class. In contrast, the \textit{Ragdoll} breed records the lowest accuracy at \textbf{79\%}, suggesting comparatively higher intra-class variability or visual similarity with other breeds.

The confusion matrix shown in Figure~\ref{fig:conf_matrix} provides a detailed overview of the model's classification performance across all the cat breeds. It highlights that the majority of predictions fall along the diagonal, indicating correct classification. Misclassifications are minimal and primarily occur between visually similar breeds reflecting the inherent challenges in fine-grained recognition tasks.

\begin{figure}[ht]
    \centering
    \includegraphics[width=1\linewidth]{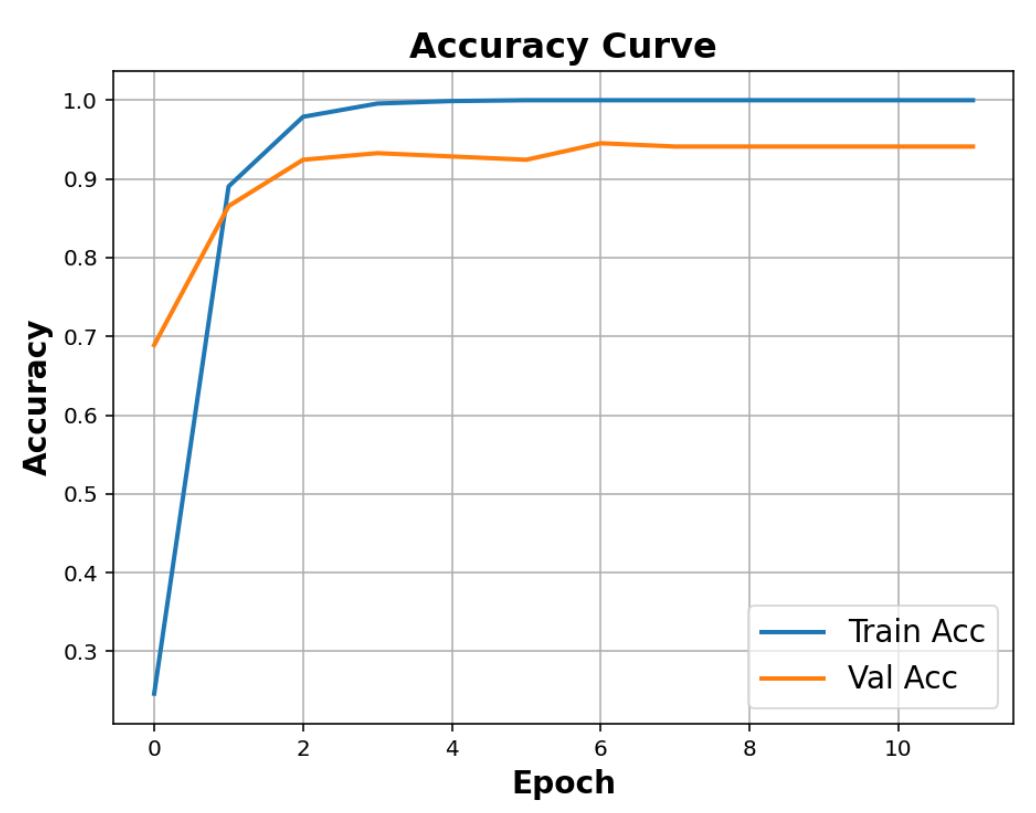}
        \caption{Training and validation accuracy over epochs}
    \label{fig:accuracy_curve}
\end{figure}

\begin{figure}[ht]
    \centering
    \includegraphics[width=1\linewidth]{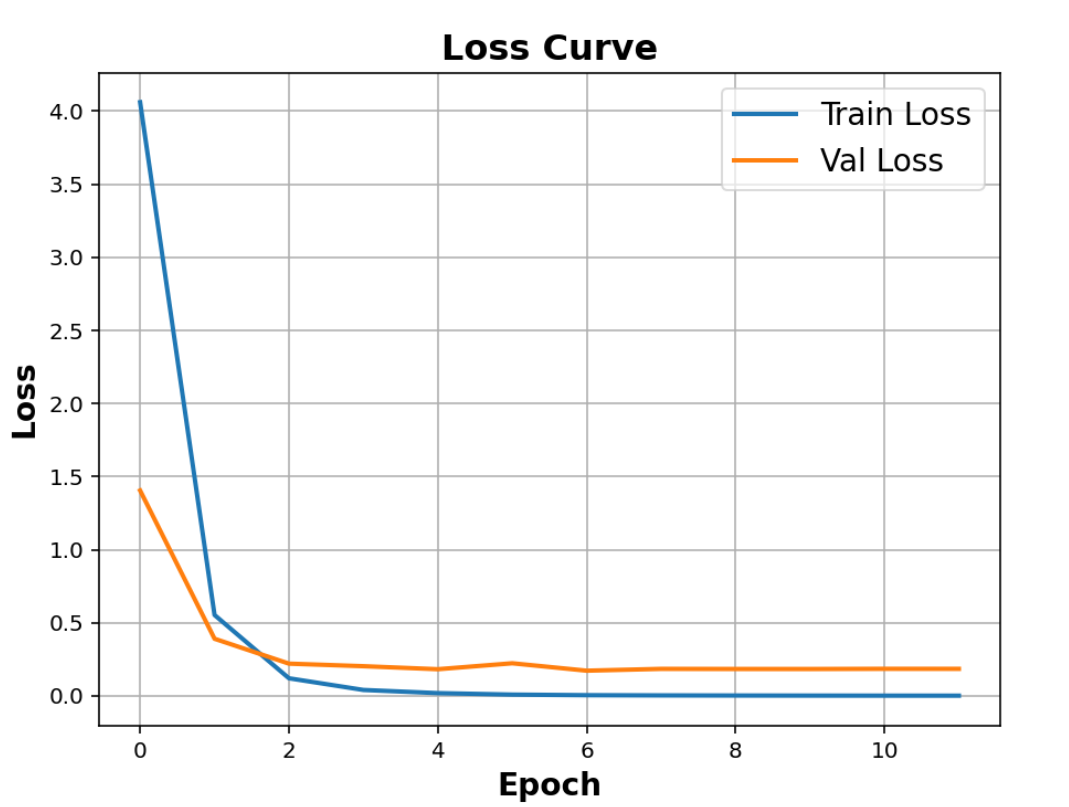}
        \caption{Loss curve over epochs}
    \label{fig:loss_curve}
\end{figure}

\begin{figure}[ht]
    \centering
    \includegraphics[width=1\linewidth]{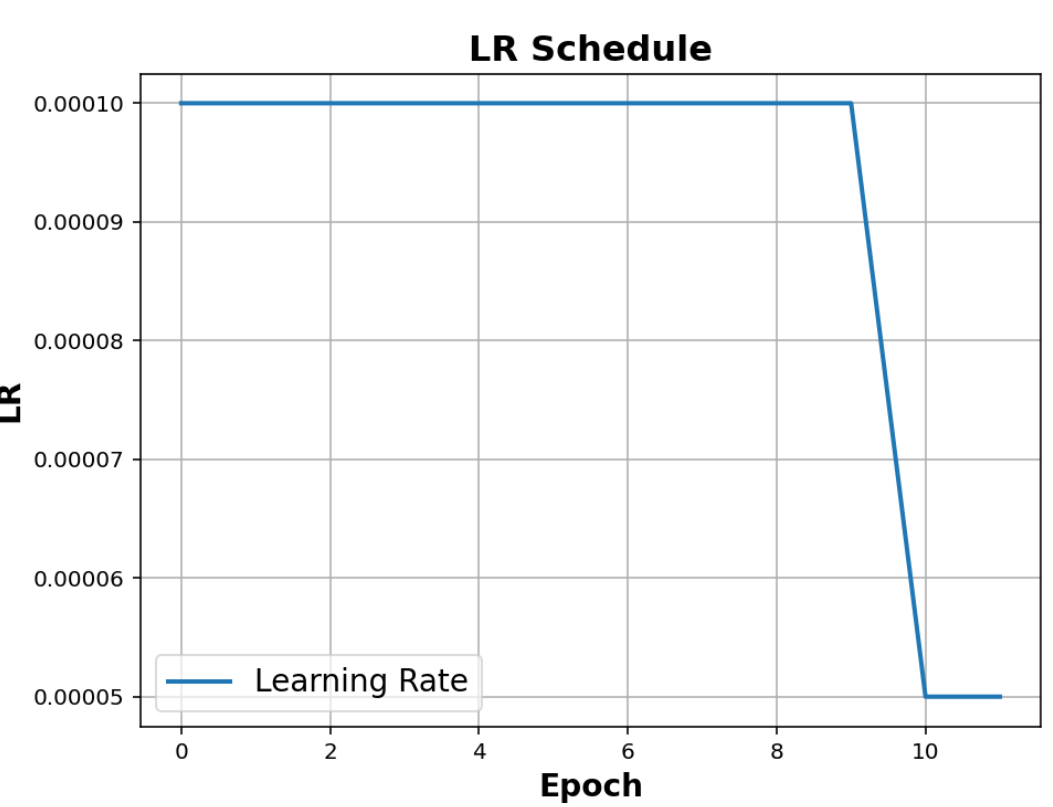}
        \caption{Learning rate over epochs}
    \label{fig:learning_rate}
\end{figure}

Figure~\ref{fig:accuracy_curve}, Figure~\ref{fig:loss_curve}, and Figure~\ref{fig:learning_rate} illustrate the training dynamics of our model across 100 epochs with a batch size of 32, an initial learning rate of $1\mathrm {e}{-4}$
, weight decay of $1 \mathrm {e}^{-4}$, and early stopping patience set to 5 epochs.

In Figure~\ref{fig:accuracy_curve}, the training accuracy rises steadily and approaches nearly $100\%$, while the validation accuracy stabilizes around $95\%$. This demonstrates that the model achieves effective learning without signs of overfitting. The smooth convergence of both curves highlights the stability of the training process under the vision transformer architecture.

Figure~\ref{fig:loss_curve} presents the loss curves for both training and validation sets. The training loss decreases rapidly during the early epochs and approaches near zero, while the validation loss converges to a small positive value. The close alignment of these curves further confirms the model’s strong generalization ability, as no significant gap is observed between training and validation losses.

Finally, Figure~\ref{fig:learning_rate} depicts the learning rate schedule employed during training. A step-wise decay strategy was applied, where the learning rate started at $1 \times 10^{-4}$ and was reduced at predefined intervals. This strategy allowed rapid progress during initial epochs while ensuring finer weight updates in later stages. The gradual decay of the learning rate contributed to stable convergence, preventing overshooting near minima and enabling the model to refine its weights effectively.

Overall, the combination of a carefully chosen batch size, learning rate schedule, weight decay, and early stopping criterion facilitated efficient training, yielding strong generalization and stable convergence.

\begin{table}[th!]
\caption{Comparison of our method with existing cat breed classification methods}
\label{tab:comparison}
\centering
\renewcommand{\arraystretch}{1.5}
\begin{tabular}{|p{2cm}|p{2.2cm}|p{2cm}|p{1.3cm}|}
\hline
\textbf{Author / Year} & \textbf{Model} & \textbf{Dataset} & \textbf{Accuracy} \\
\hline
Parkhi et al.~\cite{b2} (2012) & SIFT + HOG + Color Features & Oxford-IIIT Pet &  66.07\% \\
\hline
 Fawwaz et al. \cite{fawwaz2021klasifikasi} (2021) & VGG16 & Oxford-IIIT Pet & 60.85\% \\
\hline
Fawwaz et al. \cite{fawwaz2021klasifikasi} (2021) & InceptionV3 & Oxford-IIIT Pet & 84.94\% \\
\hline
Fawwaz et al. \cite{fawwaz2021klasifikasi} (2021) & ResNet50 & Oxford-IIIT Pet & 71.39\% \\
\hline

Cahyo et al. \cite{cahyo2023transfer} (2023) & Xception+transfer learning  & Oxford-IIIT Pet & 88.8\% \\
\hline
Ang et al. \cite {ang2024petvision}  (2024) &  MobileViTv3 (PetVision) & Oxford-IIIT Pet  & 75\% \\
\hline
\textbf{Ours (2025)} & GCViT-Tiny (Transformer) & Oxford-IIIT Pet & \textbf{92.00\%} \\
\hline

\end{tabular}
\end{table}

% \subsection*{Comparison with Existing Methods}
To contextualize the performance of our GCViT models, we compared classification accuracy with several existing methods applied to the Oxford-IIIT Pet Dataset. 
Table~\ref{tab:comparison} summarizes the reported accuracies from prior studies alongside our results. 
Our GCViT-Tiny model achieved the highest test accuracy of \textbf{92.00\%}.
substantially surpassing other methods in the table. In sepecific, Based on comparative analysis, the proposed GCViT-Tiny (Transformer) model achieves an accuracy of 92.00\% on the Oxford-IIIT Pet dataset, outperforming all previous methods. This result significantly surpasses earlier CNN methods such as VGG16 (60.85\%), InceptionV3 (84.94\%), and ResNet50 (71.39\%), and also outperforms the recently proposed hybrid model MobileViTv3 (75\%). Notably, this model also outperforms the finely tuned Xception model (88.8\%), further demonstrating the powerful representational capabilities of global attention transformers in fine-grained cat breed classification tasks. It is worth noting that our method was evaluated on a completely separate, previously unseen test set. In contrast, the methods reported in the comparison table first combined the original training, validation, and test samples, and then randomly split the dataset into new training, validation, and test subsets according to a predefined ratio before evaluation. Thus, their test data was drawn from the same distribution as their training data, whereas our evaluation was performed on genuinely unseen samples without re-accumulation or random splitting.

\section{Conclusion}
\label{sec:conclusion}
In this paper, we have presented a GCViT-based approach for fine-grained cat breed classification. The GCViT-Tiny model achieved a test accuracy of 92.00\% on the cat subset of the Oxford-IIIT Pet dataset, outperforming earlier CNN architectures and recent hybrid models. This performance gain stems from GCViT’s ability to capture both fine-grained local patterns (e.g., fur textures, facial markings) and broader contextual cues (e.g., ear shape, head proportion) within a single architecture. Unlike CNNs with limited receptive fields, GCViT’s hybrid attention mechanism offers a more comprehensive feature representation, improving robustness to variations in pose, background, and lighting. These results demonstrate that even a lightweight model can deliver strong performance when leveraging global context, making GCViT-Tiny a promising candidate for fine-grained visual recognition tasks and suitable for deployment in resource-constrained or real-time applications. Future work could explore multi-modal approaches combining visual and textual metadata, as well as extending the framework to recognize mixed-breed cats and other animal species.

\bibliographystyle{IEEEtran}
\bibliography{Reference}

@inproceedings{b2,
  author = {Parkhi, O. M. and Vedaldi, A. and Zisserman, A. and Jawahar, C. V.},
  year = {2012},
  title = {Cats and Dogs},
  booktitle = {Proceedings of the IEEE Conference on Computer Vision and Pattern Recognition (CVPR)},
  pages = {3498--3505}
}

@article{b4,
  title={Coarse-to-fine description for fine-grained visual categorization},
  author={Yao, Hantao and Zhang, Shiliang and Zhang, Yongdong and Li, Jintao and Tian, Qi},
  journal={IEEE Transactions on Image Processing},
  volume={25},
  number={10},
  pages={4858--4872},
  year={2016},
  publisher={IEEE}
}

@inproceedings{b9,
  author = {Liu, Z. and Lin, Y. and Cao, Y. and Hu, H. and Wei, Y. and Zhang, Z. and Lin, S. and Guo, B.},
  pages={10012--10022},
  year = {2021},
  title = {Swin Transformer: Hierarchical vision transformer using shifted windows},
  booktitle = {Proceedings of the IEEE International Conference on Computer Vision (ICCV)}
}

@inproceedings{b10,
  author = {Dosovitskiy, A. and Beyer, L. and Kolesnikov, A. and Weissenborn, D. and Zhai, X. and Unterthiner, T. and Dehghani, M. and Minderer, M. and Heigold, G. and Gelly, S. and Uszkoreit, J. and Houlsby, N.},
  year = {2021},
  title = {An image is worth 16x16 words: Transformers for image recognition at scale},
  booktitle = {International Conference on Learning Representations (ICLR)}
}

@inproceedings{ang2024petvision,
  title={PetVision: Improved MobileViTv3-Based Image Classification for Cat Breed Identification},
  author={Ang, Jackson and Tannaris, Shannie and Chowanda, Andry and Anderies, Anderies},
  booktitle={2024 5th International Conference on Artificial Intelligence and Data Sciences (AiDAS)},
  pages={452--456},
  year={2024},
  organization={IEEE}
}

@inproceedings{cahyo2023transfer,
  title={Transfer learning and fine-tuning effect analysis on classification of cat breeds using a convolutional neural network},
  author={Cahyo, D Diffran Nur and Sunyoto, Andi and Ariatmanto, Dhani},
  booktitle={2023 6th International Conference on Information and Communications Technology (ICOIACT)},
  pages={488--493},
  year={2023},
  organization={IEEE}
}

@article{fawwaz2021klasifikasi,
  title={Klasifikasi Ras pada Kucing menggunakan Algoritma Convolutional Neural Network (CNN)},
  author={Fawwaz, Muhammad Afif Amanullah and Ramadhani, Kurniawan Nur and Sthevanie, Febryanti},
  journal={eProceedings of Engineering},
  volume={8},
  number={1},
  pages={715--730},
  year={2021}
}

@inproceedings{hatamizadeh2023global,
  title={Global context vision transformers},
  author={Hatamizadeh, Ali and Yin, Hongxu and Heinrich, Greg and Kautz, Jan and Molchanov, Pavlo},
  booktitle={International Conference on Machine Learning},
  pages={12633--12646},
  year={2023},
  organization={PMLR}
}

\end{document}